\documentclass{article}

\usepackage{amsmath,amsfonts,bm}

\def\eqref#1{equation~\ref{#1}}

\def\1{\bm{1}}

\def\eps{{\epsilon}}

\DeclareMathAlphabet{\mathsfit}{\encodingdefault}{\sfdefault}{m}{sl}
\SetMathAlphabet{\mathsfit}{bold}{\encodingdefault}{\sfdefault}{bx}{n}

\def\tZ{{\tens{Z}}}

\newcommand{\E}{\mathbb{E}}

\newcommand{\R}{\mathbb{R}}

\usepackage{natbib}
\usepackage{authblk}
\usepackage{hyperref}
\usepackage{url}

\usepackage{microtype}
\usepackage{graphicx}
\usepackage{subfigure}
\usepackage{booktabs} 

\usepackage[utf8]{inputenc}
\usepackage{amsthm}
\usepackage{amsmath,amssymb}
\usepackage{pifont}
\usepackage{algorithm}
\usepackage{float}
\usepackage{algpseudocode}
\usepackage[margin=1in]{geometry}
\usepackage{authblk}
\usepackage{multirow}
\usepackage{wrapfig}
\usepackage{caption}
\usepackage{sidecap}

\theoremstyle{plain}
\theoremstyle{definition}
\newtheorem{definition}{Definition}[section]
\newtheorem{theorem}{Theorem}[section]
\newtheorem{proposition}{Proposition}[section]

\newtheorem{lemma}{Lemma}[section]
\newtheorem{remark}{Remark}[section]

\def\tg{{\tilde{g}}}

\renewcommand{\R}{\mathbb{R}} 

\newcommand{\cL}{\mathcal L}

\newcommand{\cN}{\mathcal N}

\newcommand{\inpro}[2]{\left\langle #1,#2 \right\rangle} 

\newcommand{\norm}[1]{\left\lVert #1\right\rVert} 

\renewcommand{\epsilon}{\varepsilon}
\renewcommand{\eps}{\epsilon}

  \newcommand{\beq}{\begin{equation}}
  \newcommand{\eeq}{\end{equation}}
  \newcommand{\beqn}{\begin{equation*}}
  \newcommand{\eeqn}{\end{equation*}}
  \newcommand{\beqr}{\begin{eqnarray}}
  \newcommand{\eeqr}{\end{eqnarray}}
  \newcommand{\beqrn}{\begin{eqnarray*}}
  \newcommand{\eeqrn}{\end{eqnarray*}}
  \newcommand{\bmline}{\begin{multline}}
  \newcommand{\emline}{\end{multline}}
  \newcommand{\bmlinen}{\begin{multline*}}
  \newcommand{\emlinen}{\end{multline*}}

\usepackage{todonotes}

\newcounter{mynotes}
\def\notes{1}
\newcommand{\gnote}[1]{\ifnum\notes=1{{\sf\color{blue} [Gopi: #1]}}\fi}
\newcommand{\jnote}[1]{\ifnum\notes=1{{\sf\color{red} [Jana: #1]}}\fi}
\newcommand{\wnote}[1]{\ifnum\notes=1{{\sf\color{green} [Woody: #1]}}\fi}

\newcommand{\grad}{\nabla}
\newcommand{\Id}{\mathrm{Id}}

\renewcommand{\tZ}{\widetilde{Z}}
\newcommand{\tz}{\widetilde{z}}

\newcommand{\vjp}{\texttt{vjp}}
\newcommand{\jvp}{\texttt{jvp}}
\graphicspath{{Figures/}}

\newcommand*{\email}[1]{\texttt{#1}}

\title{Fast and Memory Efficient Differentially Private-SGD via JL Projections\footnote{Author ordering is alphabetical. Work done when the first and the last two authors were interns at Algorithms group, Microsoft Research Redmond. }}

\date{}

\author[1]{Zhiqi Bu}
\author[2]{Sivakanth Gopi}
\author[2]{Janardhan Kulkarni}
\author[3]{Yin Tat Lee}
\author[4]{Judy Hanwen Shen}
\author[5]{Uthaipon Tantipongpipat}
\affil[1]{University of Pennsylvania\\
\email{zbu@sas.upenn.edu}}
\affil[2]{Microsoft Research\\
\email{\{sigopi,jakul\}@microsoft.com}}
\affil[3]{University of Washington\\
\email{yintat@uw.edu}}
\affil[4]{Stanford University\\
\email{jhshen@stanford.edu}}
\affil[5]{Twitter\\
\email{uthaipon@gmail.com}}

\begin{document}

\maketitle

\begin{abstract}
Differentially Private-SGD (DP-SGD) of~\cite{abadi2016deep} and its variations are the only known algorithms for private training of large scale neural networks. This algorithm requires computation of per-sample gradients norms which is extremely slow and memory intensive in practice. In this paper, we present a new framework to design differentially private optimizers called DP-SGD-JL and DP-Adam-JL. Our approach uses Johnson-Lindenstrauss (JL) projections to quickly approximate the per-sample gradient norms without exactly computing them, thus making the training time and memory requirements of our optimizers closer to that of their non-DP versions.

Unlike previous attempts to make DP-SGD faster which work only on a subset of network architectures or use compiler techniques, we propose an algorithmic solution which works for \emph{any} network in a \emph{black-box} manner which is the main contribution of this paper. To illustrate this, on IMDb dataset, we train a Recurrent Neural Network (RNN) to achieve good privacy-vs-accuracy tradeoff, while being significantly faster than DP-SGD and with a similar memory footprint as non-private SGD. 
The privacy analysis of our algorithms is more involved than DP-SGD, we use the recently proposed $f$-DP framework of~\cite{DongRS19} to prove privacy. 


\end{abstract}

\newpage
\tableofcontents
\newpage
\section{Introduction}

Over the past decade, machine learning algorithms based on (deep) neural architectures have lead to a revolution in applications such as computer vision, speech recognition and natural language processing (NLP).
An important factor contributing to this success is the abundance of data.
For most of these applications, however, the training data comes from individuals, often containing personal and sensitive information about them.
For example, natural language models for applications such as suggested replies for e-mails and dialog systems rely on
the training of neural networks on email data of users \cite{ChenLBCZLTWDCSW19, DebB19}, who may be left vulnerable if personal information is revealed. 
This could happen, for example, when a model generates a sentence or predicts a word that can potentially reveal private information of
users in the training set.
Many studies have shown successful membership inference attacks on deep learning models \cite{shokri2017membership,CarliniL19}.
Indeed, in a recent work, \cite{CarliniL19} show that ``unintended memorization" in neural networks is both commonplace and hard to prevent. 
Such memorization is not due to overtraining \cite{TetkoLL95,CarliniL19}, and ad hoc techniques such as early-stopping, dropout etc., do not prevent  the risk of privacy violations. 
Moreover, \cite{Feldman20} shows that memorization is in fact \emph{necessary}, provably, for some learning tasks.
Thus, to prevent unintended privacy breaches one needs a principled approach for private training of deep learning models. In this paper we study training neural networks with differential privacy, a mathematically rigorous notion of privacy introduced in the seminal work of \cite{DMNS06}, and focus on user level privacy.
\begin{definition}[$(\eps,\delta)$-DP]
	We say that an algorithm $M$ is $(\eps,\delta)$-DP if for any two neighboring databases $D,D'$ and any subset $S$ of outputs, we have $\Pr[M(D)\in S] \le e^\eps\Pr[M(D')\in S]+\delta.$
\end{definition}
Besides being a {\em provable} privacy notion, it has been shown that deep learning models trained with DP protect against leakage of sensitive information; we refer the readers to \cite{CarliniL19, abadi2016deep} for more details.

In a highly influential paper, ~\cite{abadi2016deep} introduced a differentially private version of stochastic gradient descent (DP-SGD) for training deep learning models, and showed that it is possible to achieve reasonable accuracy-vs-privacy tradeoff on common benchmarks such as MNIST and CIFAR10. 
Since then, there has been a vast body of work building on and extending the algorithm of~\cite{abadi2016deep}; we refer the readers to \cite{McMahan18, BuDLS19, CarliniL19, Thakkar19, Augenstein20, Zhou2020, chen2020, Balle2020}. The DP-SGD and its variations such as DP-Adam differ from their non-private counter parts in two crucial ways:

\begin{itemize}
\item \textbf{Gradient Clipping:} In each iteration of DP-SGD, we clip {\em each} per-sample gradient to have $\ell_2$-norm at most some fixed parameter $C$.
This step ensures that the {\em sensitivity} of the average gradient is bounded, which is crucial for privacy analysis. Computing the norms of per-sample gradients is the most expensive step in the algorithm. Efficient implementations of backpropagation such as in TensorFlow and PyTorch only maintain the average of per-sample gradients across a batch by default. Therefore, getting the norm of each per-sample gradient requires significantly more time and memory.
\item \textbf{Adding Noise:} Once clipped gradients are averaged across a batch, DP-SGD algorithm adds carefully calibrated noise, typically sampled from the Gaussian distribution, to ensure privacy.
\end{itemize}  
The analysis of DP-SGD in ~\cite{abadi2016deep} then follows from a careful tracking of privacy budget lost in each iteration, for which they introduced a novel technique called {\em moments accountant}, which was later generalized as Renyi differential privacy by~\cite{Mironov17_Renyi}.  This analysis was further refined and improved using the $f$-DP framework by \cite{DongRS19} and \cite{BuDLS19}, which lead to better privacy bounds. In this work, we use $f$-DP framework of \cite{DongRS19} for our privacy analysis. 

While DP-SGD has been shown to  achieve reasonable accuracy-vs-privacy tradeoff \cite{BuDLS19, abadi2016deep}, and arguably is the only known algorithm for training deep neural networks, its use in real-world deep learning has been rather limited. 
One of the primary reasons for this is the training time of DP-SGD compared to SGD.
In DP-SGD, per-sample gradients are computed at a heavy cost in runtime, especially for large deep learning models. 
The naive approach of setting the batch size to 1 is too slow to be practical as we completely lose the benefits of parallelization.
This problem has attracted significant attention in the community and has been noted in  popular implementations of DP-SGD including \cite{privacygithubTF} and \cite{privacygithubPT} (Pytorch DP).
Many strategies have been proposed to circumvent this issue, and they fall broadly into the following categories:
\begin{itemize}
\item \textbf{Microbatching:} DP-SGD implementation in Tensorflow Privacy allows dividing a batch into several microbatches and clipping the gradient at the microbatch level; the per-sample gradients in a microbatch are first averaged and then clipped, and finally these clipped gradients are averaged across microbatches. Thus, if each microbatch is of size $L$, then it gives a speedup of $L$ over the usual DP-SGD. 
Unfortunately, the sensitivity goes up by a factor of $L$, so we need to add $L$ times more noise. In our experiments, we observe that this often leads to poor accuracy even for moderate values of $L$.

\item \textbf{Multiple method:} In this approach, proposed by \cite{goodfellow} and implemented as the vectorized DP-SGD in \cite{privacygithubTF}, one copies the model as many times as there are samples in a batch. As each copy of the model is used on only one example, we can compute the per-sample gradients in parallel.
This approach improves speed at the cost of memory and is impractical for large models. 
\item \textbf{Outer product method:} This strategy was proposed by \cite{goodfellow2015efficient} for fully-connected networks and later generalized by \cite{rochette2019efficient} to include convolutional layers. The norms of per-sample gradients are computed {\em exactly} using outer products between the activations and backpropagated gradients across adjacent layers. 
In a very recent work, \cite{LK21} showed how to extend this approach to recurrent layers.
A drawback of this approach is that it does not work for all network architectures in a black-box manner, and needs careful implementation for each network architecture.
Furthermore, the current implementations of these methods require significantly more memory than non-private SGD as shown in \cite{SVK20_JAX}. 

\item \textbf{Compiler Optimization:} A completely different approach towards mitigating the problem was suggested by \cite{SVK20_JAX}.
They showed that by exploiting language primitives, such as vectorization, just-in-time compilation, and
static graph optimization, one can implement DP-SGD algorithm significantly faster.
They demonstrated these ideas in  two  frameworks: JAX and TensorFlow.
While we believe this is exciting progress, the ideas in \cite{SVK20_JAX} are  specific to these JAX and TensorFlow implementations (as of today) and present a non-algorithmic approach to this problem.
\end{itemize}

\begin{table}[!htb]
	\centering
		\caption{Summary of different methods for DP training of neural networks}

	\label{tab:Summary_DPSGD_methods}
	\begin{tabular}{|c|c|c|c|c|c|}
		\hline Optimizers & Privacy & Speed & Memory & Generalizability\\
		\hline Non-DP SGD & \ding{56} & \ding{52} & \ding{52} & \ding{52} \\
		\hline DP-SGD-Vanilla & \ding{52} & \ding{56} & \ding{52} & \ding{52} \\
		\hline DP-SGD-Multiple & \ding{52} & \ding{52} & \ding{56} & \ding{52} \\
		\hline DP-SGD-Outer & \ding{52} & \ding{52} & \ding{56} & \ding{56} \\
		\hline JAX & \ding{52} & \ding{52} & \ding{52} & \ding{56} \\
		\hline DP-SGD-JL & \ding{52} & \ding{52} & \ding{52} & \ding{52} \\
		\hline
	\end{tabular}
\end{table}

As summarized in Table~\ref{tab:Summary_DPSGD_methods}, none of these approaches for speeding up DP-SGD completely solve the problem and fall short in at least one dimension.
In this work, we propose a new {\em algorithmic framework} based on JL-projections for {\em fast and memory efficient} differentially private training of deep neural networks, which bypasses the expensive step of exactly computing per-sample gradient norms.

\subsection{Our Contributions and Techniques} 
The main idea of our algorithm is to \emph{approximate} the per-sample gradient norms instead of computing them exactly. Johnson–Lindenstrauss (JL) projections provide a convenient way to approximate the $\ell_2$ norm of a vector; simply project the vector onto a uniformly random direction, the length of the projection (scaled appropriately) is a good approximation to the $\ell_2$-norm of the vector. By doing more such projections and averaging, we get even better approximation to the true $\ell_2$-norm. Moreover, there is an efficient way to obtain such projections using forward-mode auto-differentiation or Jacobian-vector product (\jvp) (see Section~\ref{sec:jvp} for details). \jvp{} can be calculated during the forward pass making it very efficient. Since this makes the {\em sensitivity} itself a random variable, the privacy analysis is significantly harder than the traditional DP-SGD. We use the recently proposed $f$-DP framework of~\cite{DongRS19} for our analysis. Intuitively, $f$-DP captures the entire collection of $(\eps,\delta)$-DP guarantees that each iteration of the algorithm satisfies and composes them optimally to find the best $(\eps,\delta)$-DP guarantee for the final algorithm.

\sidecaptionvpos{figure}{c}
\begin{SCfigure}
\includegraphics[width=0.5\textwidth]{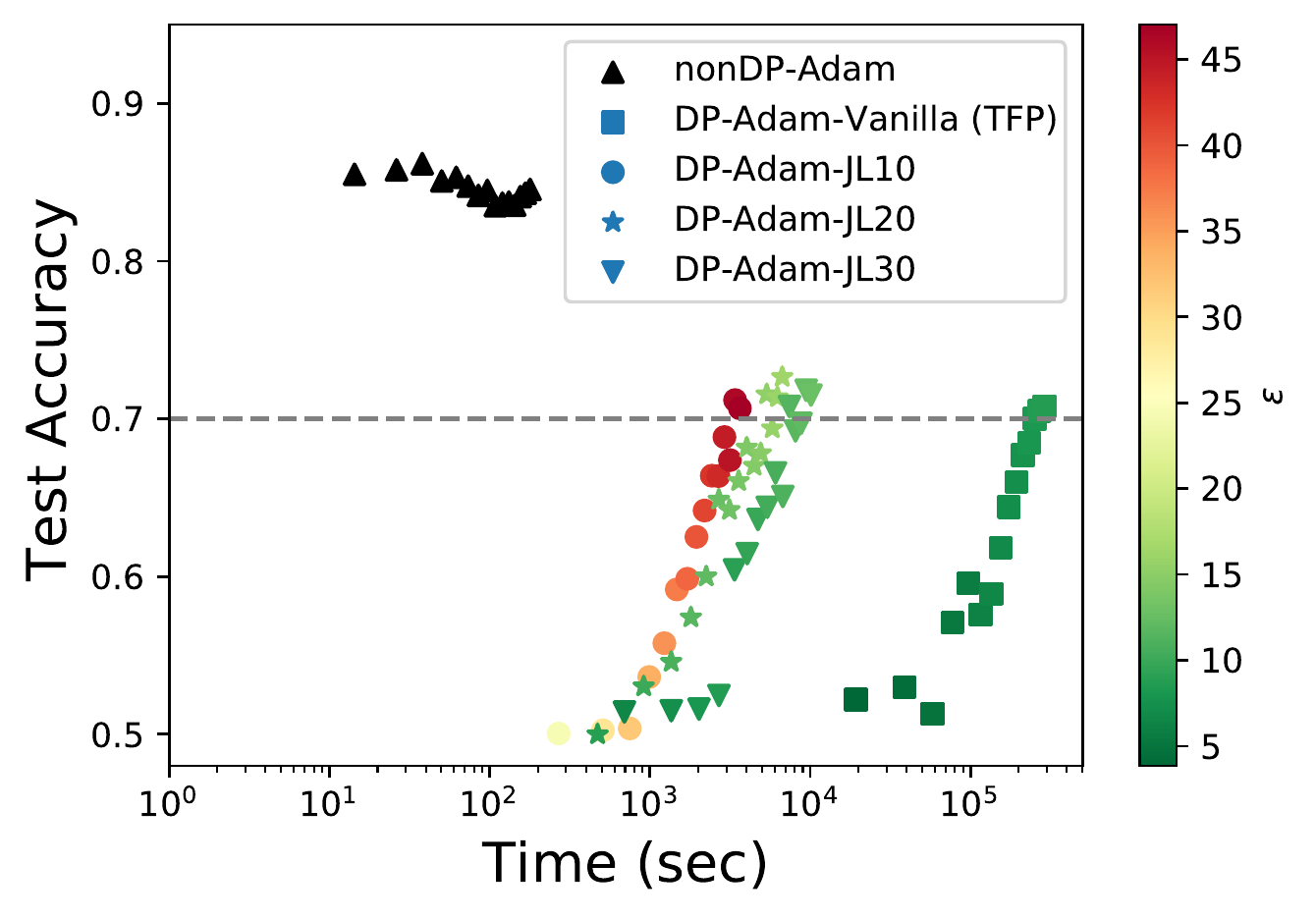}
\caption{Performance of various algorithms training an RNN on IMDb dataset using a batch size of 256. The privacy parameter $\epsilon$ is color coded for $\delta=10^{-5}$. In DP-Adam-JL$r$, $r$ refers to the JL dimension, i.e., the number of JL projections used to approximate per-sample gradient norms. 
DP-Adam-Vanilla (TFP) is the implementation of DP-Adam in \cite{privacygithubTF} library.} 
\label{fig:RNNcolorplot}
\end{SCfigure}

To summarize, the key contributions of this paper are:
\begin{itemize}
\item Our algorithms DP-SGD-JL and DP-Adam-JL are considerably faster than previously known differentially private training algorithms that require exact per sample gradient norms, and work for {\em all} network architectures. The privacy-vs-accuracy tradeoff achieved by our algorithms is comparable to the existing state-of-the-art DP-algorithms.

\item Memory footprint of our algorithms is nearly the same as that of non-private training algorithms. This allows us to work with larger batch sizes (which is crucial for achieving good privacy-vs-accuracy tradeoffs) without resorting to gradient accumulation. This also improves the running time of training. 

\item Compared to DP-SGD, our analysis of privacy is more involved. Since we only approximate the per-sample gradient norms, we cannot precisely bound {\em sensitivity}. Therefore the analysis requires significantly new ideas, which could be of independent interest.

\item We demonstrate these improvements by training an RNN using layers such as bidirectional LSTM, embedding layer, fully connected etc. on the IMDb dataset.
As can be seen from Figure \ref{fig:RNNcolorplot}, our algorithms are significantly faster than current implementations of DP-SGD while achieving similar privacy-vs-accuracy tradeoff. 

\item Our algorithms introduce a new \emph{knob}, the dimension of JL-projection,  which allows us to do a tradeoff between training time and privacy, which was not possible in earlier algorithms.
All hyperparameters being the same, smaller JL dimension will give much better running time with an increase in privacy budget.
Moreover, our experiments show that although the privacy bounds we could prove for DP-SGD-JL are not so great for very small JL dimensions (see Figure~\ref{fig:RNNprivacy}), their behavior (accuracy-vs-epoch curve) converges very quickly to that of DP-SGD-Vanilla. Figure~\ref{fig:MNISTHyper} shows that DP-SGD-JL(3) is already very close to DP-SGD-Vanilla and DP-SGD-JL(20) is almost indistinguishable.
We find these properties of DP-SGD-JL algorithms to be very useful during initial stages of experimentation and hyper-parameter tuning for private training.

\end{itemize}


\section{DP-SGD-JL Algorithm}
\label{sec:algorithm}
In this section we describe our new differentially private optimizers. We will first present DP-SGD-JL, and DP-Adam-JL follows in the same lines and is presented in Appendix~\ref{sec:DP-Adam-JL}. We begin with an introduction to ``Jacobian-vector product'' (\jvp) which is crucial for our algorithm.

\subsection{Jacobian-vector product (\jvp)}
\label{sec:jvp}
Given a function $f:\R^d \to \R$, the gradient of $f$ with respect to $\theta$, denoted by $\grad_\theta f$, is:
$$\grad_\theta f = \left(\frac{\partial f}{\partial \theta_1},\frac{\partial f}{\partial \theta_2},\dots, \frac{\partial f}{\partial \theta_d}\right).$$
Let $F:\R^d \to \R^m$ be some function given by $F(\theta)=(F_1(\theta),F_2(\theta),\dots,F_m(\theta))$. The Jacobian of $F(\theta)$, denoted by $\grad_\theta F$, is the matrix:
$$\grad_\theta F = \left[\frac{\partial F_i}{\partial \theta_j}\right]_{ij} = \begin{bmatrix} \grad_\theta F_1 \\
\grad_\theta F_2 \\
\vdots\\
\grad_\theta F_m
\end{bmatrix}.$$
Most auto-differentiation packages allow for calculating the vector-Jacobian product (\vjp) given by $u^T\grad_\theta F=\sum_{i=1}^m u_i\grad_\theta  F_i$ for any $u\in \R^m$ efficiently using reverse-mode auto-differentiation, which is the familiar `backpropagation'.\footnote{In PyTorch, $u^T{\grad_\theta F}$ can be calculated as \texttt{autograd.grad(F,$\theta$,grad\_outputs=u)}. In TensorFlow, this is \texttt{tf.GradientTape().gradient(F,$\theta$,output\_gradients=u)}.} One can also calculate the Jacobian-vector product (\jvp) given by 
$$\grad_\theta F\cdot v=\begin{bmatrix}\inpro{\grad_\theta  F_1}{v}\\ \inpro{\grad_\theta  F_2}{v}\\ \vdots \\ \inpro{\grad_\theta  F_m}{v} \end{bmatrix}$$ 
efficiently using forward-mode auto-differentiation (i.e., the required derivatives are calculated during the forward pass of the network). We refer the reader to the survey on automatic differentiation by~\cite{Baydin17automatic} for more details. \jvp{} is implemented using forward-mode auto-differentiation in the recent TensorFlow versions.\footnote{Supported in tf-nightly$\ge$2.4.0.dev20200924 as \texttt{tf.autodiff.ForwardAccumulator($\theta$,v).jvp(F)}.} Unfortunately, PyTorch doesn't have an implementation of forward-mode auto-differentiation. Instead, one can compute \jvp{} using two calls to \vjp{}, this is called the `double \vjp{} trick' (see \cite{jtown}).\footnote{In Pytorch, an implementation of \jvp{} using the double \vjp{} trick exists and can be invoked as \texttt{torch.autograd.functional.jvp(F,$\theta$,inputs=v)}.}
Define $G(\alpha)=\alpha^T \grad_\theta  F$, which can be calculated using \vjp. Note that $\grad_\alpha G = (\grad_\theta  F)^T.$ Now we can use \vjp{} again on $G$ to calculate \jvp{} as $$v^T\grad_\alpha G = v^T (\grad_\theta  F)^T = (\grad_\theta  F \cdot v)^T.$$ In our experiments, we use the efficient implementation of \jvp{} in TensorFlow to compute the Jacobian-vector products as the double $\vjp{}$ trick is a few times slower.

\subsection{Algorithm}

\begin{wrapfigure}[10]{r}{0.4\linewidth}
\centering
\vspace{-1cm}
\includegraphics[width=\linewidth]{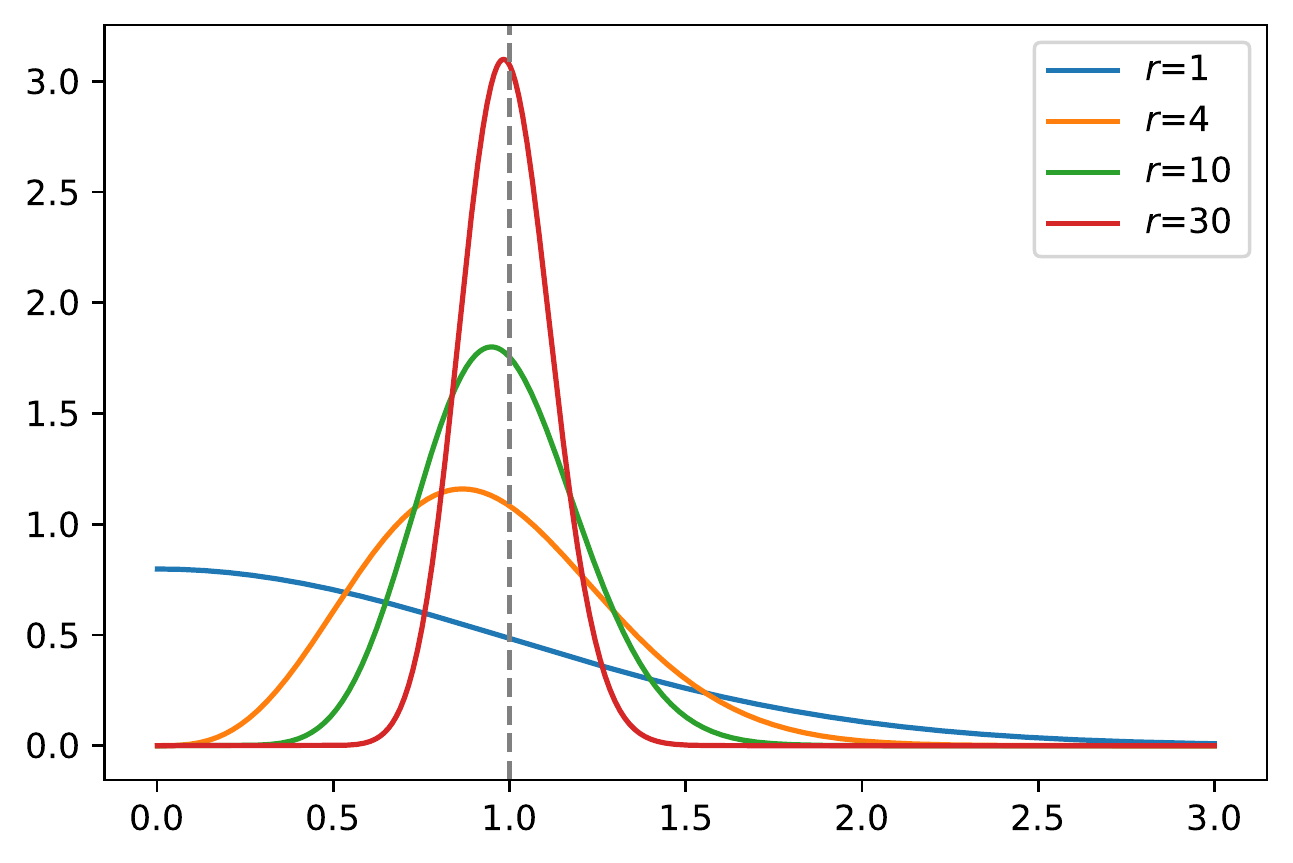}
\vspace{-0.5cm}
\caption{Distribution of $\sqrt{\frac{1}{r}\chi^2_r}$ for different $r.$}
\label{fig:JL_approximation}
\end{wrapfigure}

The main idea of our algorithm is to approximate $\ell_2$-norms of per-sample gradient norms instead of computing them exactly. And the key tool to achieve this is JL projections.
\begin{proposition}[JL projections]
   Let $y\in \R^d$ be any vector. If $v_1,v_2,\dots,v_r\sim \cN(0,I_d)$ are independent standard Gaussian vectors, then $M_r=\sqrt{\sum_{i=1}^r\frac{1}{r}\inpro{y}{v_i}^2}$ has the same distribution as $\norm{y}_2 \sqrt{\frac{1}{r}\chi^2_r}$.\footnote{$\chi^2_r$ is the chi-square distribution with $r$ degrees of freedom which is the distribution of sum of squares of $r$ standard Gaussians.} In particular $\E[M_r^2]=\norm{y}_2^2.$
\end{proposition}
\begin{proof}
   By the properties of the standard Gaussian distribution, $\inpro{y}{v_i}$ has the distribution of $\norm{y}_2\cN(0,1).$ And $\inpro{y}{v_i}$ are independent for $i=1$ to $r$. Therefore $\sum_{i=1}^r \inpro{y}{v_i}^2$ has the same distribution as $\norm{y}_2^2 \chi^2_r.$
\end{proof}

As shown in Figure~\ref{fig:JL_approximation}, as $r$ grows larger, the distribution of $\sqrt{\frac{1}{r}\chi^2_r}$ concentrates more around 1 and therefore $M_r$ becomes a better estimate of $\norm{y}_2.$ Using \jvp{}, we can compute projections of per-sample gradients on to standard Gaussian vectors quickly and therefore get good approximations to their norms. This is the main idea of DP-SGD-JL (Algorithm~\ref{alg:DPSGD_JL}). The privacy analysis of our algorithm is quite involved and is presented in Section~\ref{sec:privacy_analysis}.

\begin{algorithm*}[!ht]
\caption{Differentially private SGD using JL projections (DP-SGD-JL)}
 \label{alg:DPSGD_JL}
 \begin{algorithmic}
   \State {\bfseries Input:} Examples $\{x_1,x_2,\dots,x_N\}$, loss function $\cL(\theta)=\E_{i\in [N]}[\cL(\theta;x_i)]$, initialization $\theta_0$.\\
    Parameters: number of iterations $T$, learning rates $(\eta_1,\eta_2,\dots,\eta_T)$, noise scale $\sigma$, batch size $B$, clipping norms $(C_1,C_2,\dots,C_T)$, number of JL projections $r$.
   \State
   \For{$t=1$ to $T$}
   \State Sample $S_t=\{X_1,X_2,\dots,X_B\}\subset \{x_1,x_2,\dots,x_N\}$ uniformly at random. 
   \State Define $F(\theta)= \left(\cL(\theta;X_1),\cL(\theta;X_2),\dots,\cL(\theta;X_B)\right)$
   \State
   \State Sample $v_1,v_2,\dots,v_r \leftarrow \cN(0,I_d)$ (where $\theta\in\R^d$)
   \Comment{JL projections to estimate per-sample gradient norm}
   \For{$j=1$ to $r$}
   \State  $(P_{1j},P_{2j},\dots,P_{Bj}) \leftarrow \grad_\theta F(\theta_{t-1}) \cdot v_j$ (using \jvp)
   \Comment{Note that $P_{ij}=\inpro{\grad_\theta \cL(\theta_{t-1};X_i)}{v_j}$}
   \EndFor
   \For{$i=1$ to $B$}
   \State Set $M_i = \sqrt{\frac{1}{r}\sum_{j=1}^r P_{ij}^2}$
   \Comment{$M_i$ is an estimate for $\norm{\grad_\theta \cL(\theta_{t-1};X_i)}_2.$}
   \EndFor
   \State Define $\widetilde{\cL}(\theta)=\frac{1}{B}\sum_{i\in B} \min\{1,\frac{C_t}{M_i}\}\cdot \cL(\theta;X_i).$
  \Comment{Scale the losses to clip per-sample gradients}
   \State
   \State $\tg_t \leftarrow \grad_\theta \widetilde{\cL}(\theta_{t-1}) + \frac{\sigma \cdot C_t}{B}\cdot \cN(0,I_d)$
   \Comment{Add noise to the average of clipped gradients}
   \State
   \State Update $\theta_t\leftarrow \theta_{t-1} - \eta_t \tg_t$
   \EndFor
\State {\bfseries Output:} $\theta_0,\theta_1,\theta_2,\dots,\theta_T$
\end{algorithmic}
\end{algorithm*}

\section{Experiments}
\label{sec:experiments}

In this section, we demonstrate experimentally that compared to existing implementations of DP-SGD with exact per-sample gradient clipping, our optimizers have significant advantages in speed and memory cost while achieving comparable accuracy-vs-privacy tradeoff. Moreover our algorithms perform well on a variety of network architectures.
The main goal of this section is to give empirical evidences towards the following three strengths of our algorithm alluded in the introduction:
\begin{enumerate}
	\item Our algorithm is significantly faster compared to per-sample gradient computations and works for any network in a black-box way. 
	\item Memory footprint of our algorithm is roughly same as non-private SGD. 
	\item The DP-SGD-JL algorithms with smaller values of JL dimension exhibit similar behavior as DP-SGD but with orders of magnitude speed up, and hence can be used for hyper-parameter search.
\end{enumerate}

In the following, we write `nonDP-SGD' for the standard non-private SGD and `DP-SGD-Vanilla' for the implementation of DP-SGD in \cite{privacygithubTF}, nonDP-Adam and DP-Adam-Vanilla are similarly defined. We use Tensorflow and \cite{privacygithubTF} for all our experiments because \cite{privacygithubPT} does not support arbitrary network architectures.\footnote{In \href{github.com/pytorch/opacus/blob/master/tutorials/building_lstm_name_classifier.ipynb}{Pytorch Opacus github}, the LSTM layer is only partially supported, e.g. single directional, single LSTM layer, no dropout layer; other recurrent layers such as GRU are not supported (see \href{https://github.com/pytorch/opacus/blob/master/opacus/supported_layers_grad_samplers.py}{Opacus}).} Moreover Tensorflow has an efficient implementation of \jvp{} while PyTorch doesn't.
\paragraph{Notation:} We denote the noise multiplier as $\sigma$, clipping norm as $C$, batch size as $B$, learning rate as $\eta$ and the number of epochs as $E=BT/N$. We fix the privacy parameter $\delta=10^{-5}$, as done by prior work. We denote the JL dimension used by each optimizer in the parentheses. We use one Tesla P100 16GB GPU for all experiments. In all the experiments, we report the time per epoch by averaging over a large number of epochs.

\subsection{Training an LSTM model on IMDb dataset}
\label{sec:IMDB}
\begin{table}[!ht]
\centering
\scalebox{0.9}{
	\begin{tabular}{|c|c|c|c|c|c|}
	\hline
	Non-DP Adam&DP-Adam-Vanilla&DP-Adam-JL(1)&DP-Adam-JL(5)&DP-Adam-JL(10)&DP-Adam-JL(30)
	\\\hline
	12&19317&41&128&243&675
	\\\hline
\end{tabular}
}
\captionof{table}{Seconds per epoch to train our RNN with 598,274 parameters and 25,000 training samples. We set $\beta_1=0.9, \beta_2=0.999, \sigma=0.6, C=1, B=256, \eta=0.001, E=15$.}
\label{table:RNN embedding}
\end{table}
The goal of these experiments is to demonstrate the first strength of our algorithm: it is significantly faster than per-sample gradient computations and works for any network in a black-box way. 
We train a bidirectional LSTM with an embedding layer on the IMDb dataset for sentiment analysis. 
We remark that simpler networks {\em not} based on RNNs  can achieve good accuracy-vs-privacy tradeoff as shown in \cite{BuDLS19} and \cite{imdb_readme}.
However, LSTM models based on RNN architectures are widely used in NLP applications, and hence efficient private training of such models remains an important challenge in this area \cite{McMahan18}.
Moreover, as we noted in the introduction, extensions of outer product trick to bidirectional LSTMs as described in  \cite{LK21} are significantly more complicated, and require considerable effort to implement.
Since authors of \cite{LK21} did not provide the code, we are unable to compare the improvements. 
Moreover, as also noted in \cite{LK21}, the outer product method  requires significantly more memory and hence will not scale to large batch sizes, which is very important to achieve good privacy vs utility tradeoff.

We implement DP-Adam-JL using \texttt{jvp} method in TensorFlow. We train the same single-layer bidirectional LSTM\footnote{We cannot use CuDNNLSTM and instead use LSTM for the following reason. When using CuDNNLSTM (and on GPU), we observe a significant speedup compared to LSTM, but the accuracy is invalid and we further incur a \href{https://github.com/tensorflow/tensorflow/issues/37091}{LookupError} when computing \texttt{jvp}.} as in the \cite{rnntutorial} tutorial, using the same IMDb dataset with 8k vocabulary. 
The dataset has binary labels and is preprocessed by zero-padding the sequences to a length of 150 before being fed into the embedding layer.

\begin{wrapfigure}[17]{l}{0.5\linewidth}
\centering
\vspace{-0.5cm}
\includegraphics[width=\linewidth]{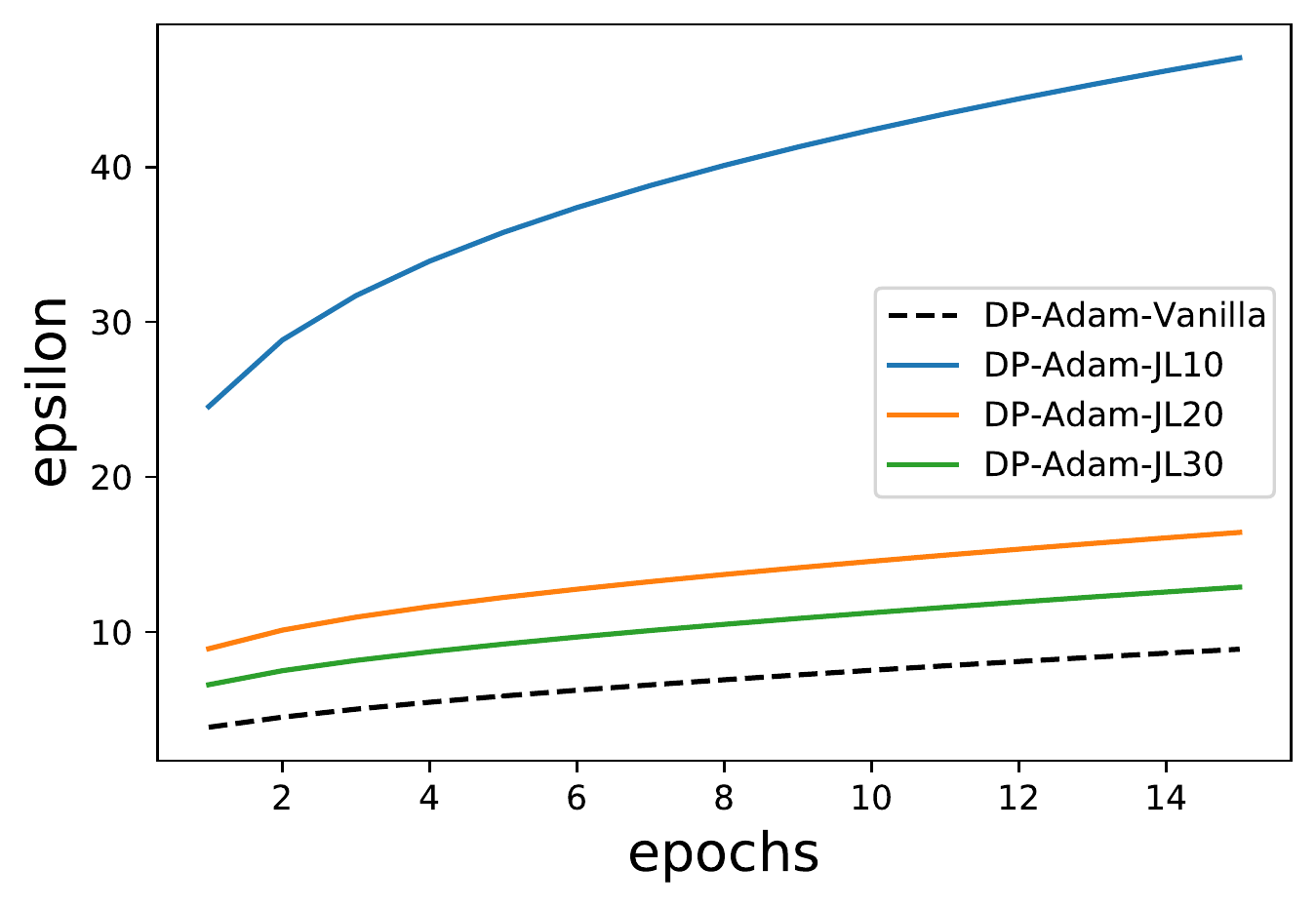}
\vspace{-0.6cm}
\caption{Privacy curves for various algorithms while training the LSTM model on IMDb dataset. Here $\delta=10^{-5}$ is fixed.}
\label{fig:RNNprivacy}
\end{wrapfigure}

Table \ref{table:RNN embedding} shows the training time per epoch for different algorithms.
As expected, we observe that DP-Adam-JL algorithms with smaller values of JL dimension are significantly faster than DP-Adam-Vanilla. 
However, as we note in the Figure~\ref{fig:RNNprivacy} privacy guarantees of DP-Adam-JL algorithm with smaller values of JL dimension are considerably worse than  DP-Adam-Vanilla.  
On the other hand, DP-Adam-JL(30) is $30\times$ faster than DP-Adam-Vanilla while achieving similar privacy guarantees as DP-Adam-Vanilla. 
When allowed to train for sufficient number of epochs, we observed that all the algorithms achieved same accuracy but with different privacy costs and running times.
This three dimensional tradeoff between utility, privacy and speed is depicted in the Figure \ref{fig:RNNcolorplot} (see introduction), where we plot the privacy values using a color plot.

\begin{remark}
As we can see from Table~\ref{table:RNN embedding}, for a batch size of 256, the slowdown of DP-Adam-Vanilla is more than 256 compared to nonDP-Adam. 
This is counter intuitive as a naive implementation DP-Adam-Vanilla, by setting the batch size equal to 1 and then doing gradient accumulation across 256 batches should only be 256 times slower.
As we show below, this is due to the memory issues of DP-Adam-Vanilla in the implementation in \cite{privacygithubTF}. Indeed the naive implementation of DP-Adam-Vanilla in tensorflow using gradient accumulation takes about 4000 seconds per epoch for the same batch size.
\end{remark}

\subsection{Memory footprint}
\label{sec:memory}
Another key strength of JL based algorithms is their memory consumption, and the goal of this section is to show this aspect of our new algorithms via experiments. 
As a proxy for memory consumption, we compare the largest batch size each algorithm can handle without running out of memory.
It is known that to achieve good privacy-vs-accuracy tradeoffs for DP-SGD, we need to use large batch sizes~\cite{abadi2016deep}.
One way to support large batch sizes is via gradient accumulation;
however, this has the disadvantage that one loses parallelism, which in turns leads to slower run times.
Hence memory footprint of algorithms also indirectly affects the training time.

We compare our JL algorithms with the implementation of DP-SGD-Vanilla in \cite{privacygithubTF}. We train a convolutional neural network from \cite{privacygithubTF} tutorial on MNIST dataset, which has 60,000 training samples.\footnote{We use the implementation and the network from \texttt{mnist\_dpsgd\_tutorial\_keras.py}}
As we can see from Table \ref{table:MNISTmemory}, DP-SGD-JL algorithm and nonDP-SGD can both run with the maximum possible batch size of 60,000 whereas DP-SGD-Vanilla can only handle a batch size of at most 500.
In general, we believe that the memory footprint of DP-SGD-JL algorithm is very close to that of non-private SGD. 
To show this, we augment the CNN in \cite{privacygithubTF} tutorial with dense layers to blowup the model size to 17,146,938 parameters, and repeat the experiment. 
As we see from Table \ref{table:MNISTmemory2}, the largest batch size supported by DPSGD-JL(30) is only a factor 2 away from the largest batch size supported by non-DP SGD. On the other hand, we observe that DP-SGD-Vanilla only supports a batch size of 100.

\begin{table}[!htb]
\centering
\scalebox{1}{
	\begin{tabular}{|c|c|c|c|}
	\hline
	nonDP-SGD&DP-SGD-Vanilla&DP-SGD-JL(10)&DP-SGD-JL(30)
	\\\hline
	60,000&512&60,000&60,000
	\\\hline
\end{tabular}
}
\captionof{table}{Maximum batch size supported by various algorithms on a CNN with 26,010 parameters trained on MNIST with 60,000 training samples. Training done on one Tesla P100 16GB GPU.}
\label{table:MNISTmemory}
\end{table}

\begin{table}[!htb]
\centering
\scalebox{1}{
	\begin{tabular}{|c|c|c|c|}
	\hline
	Non-DP SGD&DP-SGD-Vanilla&DP-SGD-JL(10)&DP-SGD-JL(30)
	\\\hline
	52,000&100&28,000&25,000
	\\\hline
\end{tabular}
}
\captionof{table}{Maximum batch size supported by various algorithms on a CNN with 17,146,938 parameters trained on MNIST with 60,000 training samples. Training done on one Tesla P100 16GB GPU.}
\label{table:MNISTmemory2}
\end{table}

The above experiments also give a possible explanation of why DP-SGD-Vanilla implementation in TFP has a slowdown that is larger than the batch size, as we observed in the LSTM experiments. Even for MNIST, we observe that DP-SGD-Vanilla running time gets better with batch size in the very beginning but as the batch size becomes larger the running time gets worse, and soon after it runs out of memory. 

\subsection{Using DP-SGD-JL for hyper-parameter search}
\begin{SCfigure}
\includegraphics[width=0.6\linewidth]{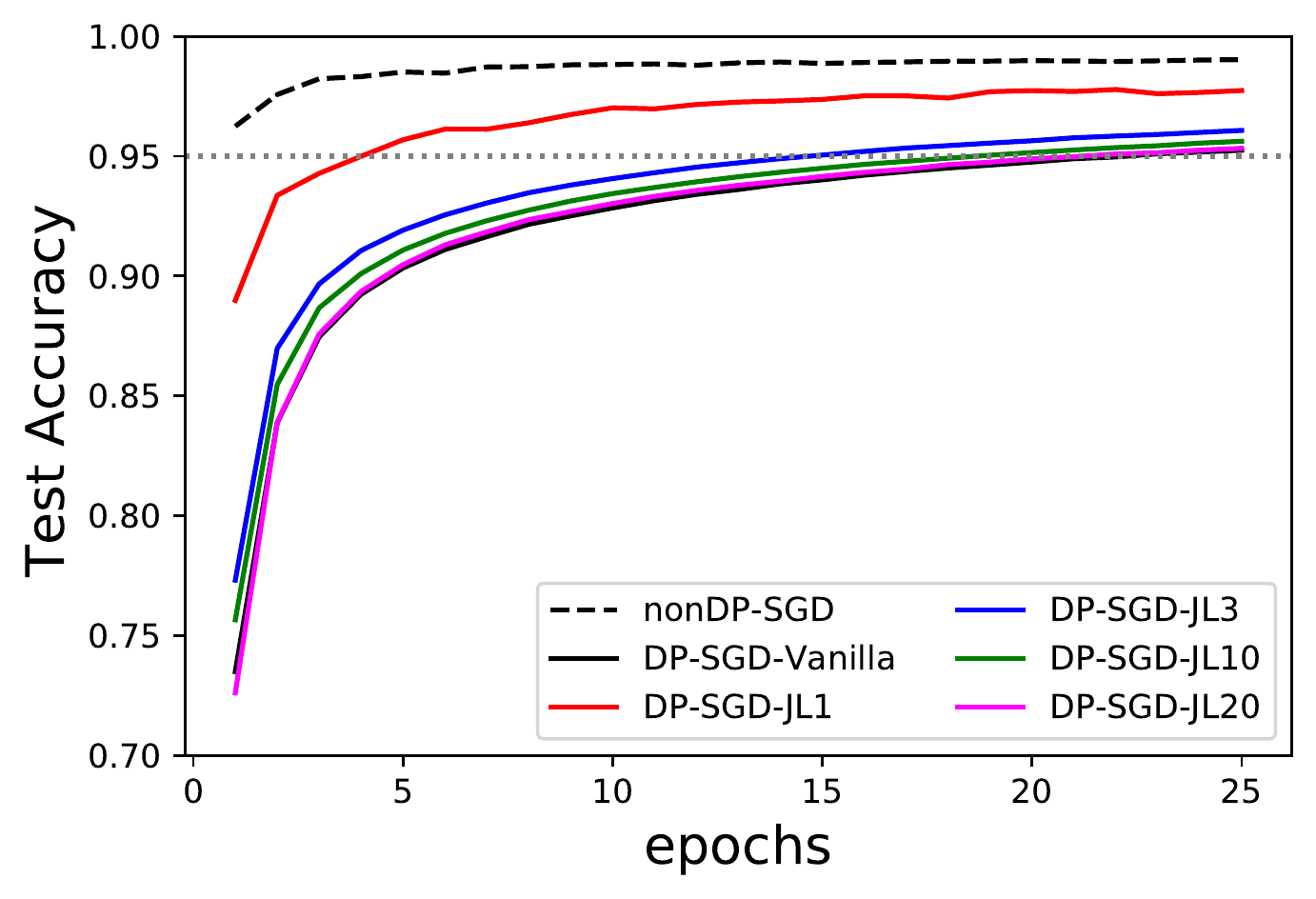}
\caption{Behavior of various private algorithms for MNIST dataset. The test accuracy is averaged over 25 runs. Note that the DP-SGD-JL20 curve is nearly overlapping with the curve for DP-SGD-Vanilla.}
\label{fig:MNISTHyper}
\end{SCfigure}
As we saw in our experiments summarized in Table \ref{table:RNN embedding}, our algorithms with small values of JL dimension are orders of magnitude faster than DP-Adam-Vanilla; DP-Adam-JL(1) is about 470x faster and DP-SGD-JL(5) is about 150x faster.
However, unfortunately, the privacy bounds we can prove for these algorithms are considerably worse than DP-Adam-Vanilla.
Despite this drawback, we observe that behavior of DP-SGD-JL even for small JL dimension is very close to that of DP-Adam-Vanilla.
Figure \ref{fig:MNISTHyper} plots the accuracy vs epochs for various algorithms training a CNN from \cite{privacygithubTF} tutorial on MNIST dataset.
We observe that as JL dimension increases, the accuracy vs epoch curve quickly converges to that of DP-SGD-Vanilla. DP-SGD-JL(3) is already very similar to DP-SGD-Vanilla and DP-SGD-JL(20) is nearly indistinguishable. This also lets us hypothesize that the privacy of our algorithms could be much better than what we could prove and that it should converge equally quickly to that of DP-SGD-Vanilla.
Thus we believe that DP-SGD-JL(3) or DP-SGD-JL(5) are good candidates for experimentation and hyper-parameter tuning during private training, since their behavior is almost identical to that of DP-SGD-Vanilla while being orders of magnitude faster.

\section{Privacy Analysis}
\label{sec:privacy_analysis}

\subsection{$f$-DP preliminaries}

We use the recently proposed $f$-DP framework of~\cite{DongRS19} for our privacy analysis. The $f$-DP framework allows us to reason about a collection of $(\eps,\delta)$-privacy guarantees simultaneously which can then be composed to get a much better $(\eps,\delta)$-privacy for the final algorithm. We will first define the notion of $(\eps,\delta)$-DP formally and then define the notion of $f$-DP. We then state a proposition from~\cite{DongRS19} which shows that these two notions are dual to each other.

\begin{definition}[$(\eps,\delta)$-DP]
	We say that an algorithm $M$ is $(\eps,\delta)$-DP if for any two neighboring databases $D,D'$ and any subset $S$ of outputs, we have $\Pr[M(D)\in S] \le e^\eps\Pr[M(D')\in S]+\delta.$
\end{definition}
For each value of $\eps\ge 0$, there exists some $\delta \in [0,1]$ such that $M$ is $(\eps,\delta)$-DP. We can represent all these privacy guarantees by a function $\delta(\eps):\R^{\ge 0} \to [0,1]$ and say that $M$ is $(\eps,\delta(\eps))$-DP for each $\eps\ge 0.$ We will now see that there is a dual way to represent the function $\delta(\eps)$ called $f$-DP. To introduce this, we will need the notion of a tradeoff function.

\begin{definition}[Tradeoff function]
	Let $P,Q$ be two distributions over some domain $X.$ Let $\phi:X \to [0,1]$ be a prediction rule which given a sample predicts which distribution the sample came from. The type I error is defined as $\alpha=\E_{x\sim P}[\phi(x)]$ and the type II error is defined as $\beta=\E_{x\sim Q} [1-\phi(x)]$. The tradeoff function $T(P||Q):[0,1]\to [0,1]$ is defined as: 
	\begin{equation}
	\label{eqn:tradeoff_curve}
	T(P||Q)(\alpha)=\inf_\phi\{1-\E_{Q} [\phi]: \E_P[\phi]\le \alpha\}.
	\end{equation}

	Given two random variables $X,Y$, we define $T(X||Y)$ to be $T(P||Q)$ where $P,Q$ are the distributions of $X,Y$ respectively.
\end{definition}
Note that if $T(P,Q)=f$, then $T(Q,P)=f^{-1}.$ A tradeoff curve $f$ is called symmetric if $f^{-1}=f.$
Given two functions $f,g$ on the same domain, we say $f\preceq g$ if $f(x)\le g(x)$ for all $x$ in the domain. $f \succeq g$ is similarly defined.


\begin{definition}[$f$-DP]
	We say an algorithm $M$ is $f$-differentially private if for every two neighboring databases $D,D'$, we have $T(M(D)||M(D'))\succeq f$.
\end{definition}

\begin{wrapfigure}[15]{l}{0.4\linewidth}
\centering
\vspace{-0.6cm}
\includegraphics[width=\linewidth]{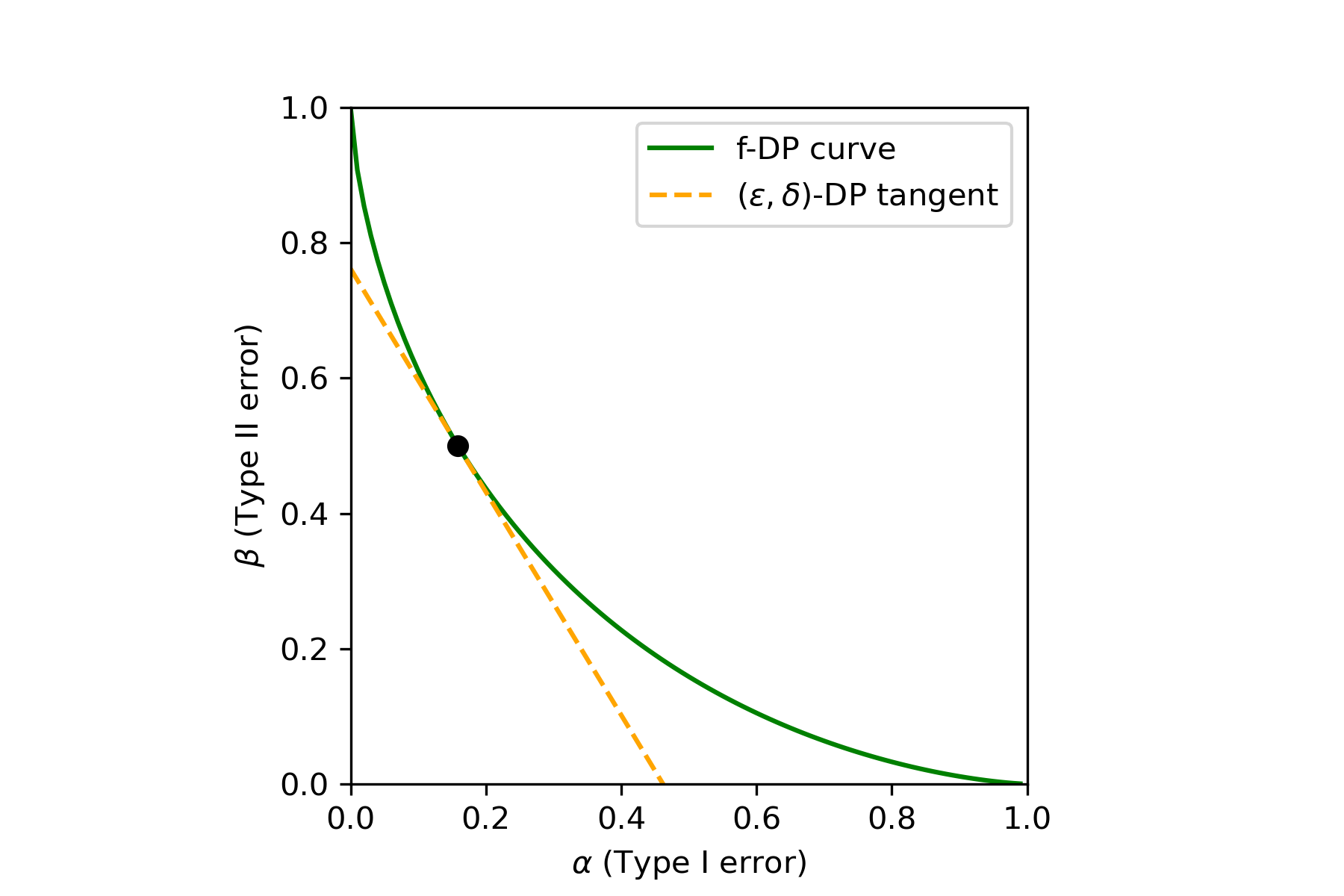}
\vspace{-0.4cm}
\caption{Duality of $f$-DP and $(\eps,\delta)$-DP. Every tangent to the $f$-DP curve gives an $(\eps,\delta)$-DP guarantee where the slope is $-e^\eps$ and $y$-intercept is $1-\delta$.}
\label{fig:fDP_eps_delta_duality}
\end{wrapfigure}

\begin{proposition}[Duality of $f$-DP and $(\eps,\delta)$-DP from~\cite{DongRS19}]
	\label{prop:fDP_epsdeltaDP}
	If $M$ satisfies $f$-DP where $f$ is symmetric (i.e., $f^{-1}=f$). Let $\alpha^*$ be such that $f(\alpha^*)=\alpha^*$. Then $M$ satisfies $(\eps(\alpha),\delta(\alpha))$-DP for every $0\le \alpha \le \alpha^*$ where:
	$$\eps(\alpha)=\log(-f'(\alpha)),\ \delta(\alpha)=1-f(\alpha)+\alpha f'(\alpha).$$ So the tangent to $f$ at $\alpha$ has slope $-\exp(\eps(\alpha))$ and $y$-intercept $1-\delta(\alpha)$ (see Figure~\ref{fig:fDP_eps_delta_duality}).
\end{proposition}

Note that by convexity of $f,$ $f':[0,1]\to \R$ is increasing in $[0,1]$ and by symmetry $f'(\alpha^*)=0.$ Therefore Proposition~\ref{prop:fDP_epsdeltaDP} covers the entire range of $\eps\ge 0.$



\begin{proposition}[Post-processing]
\label{prop:post-processing}
	Let $X,Y$ be two random variables supported on $A$ and let $M:A\to B$ is some randomized function, then $T(X||Y) \preceq T(M(X)||M(Y)).$
\end{proposition}

\begin{proposition}
\label{prop:sufficient-statistic}
	Let $(X_1,Y_1)$ and $(X_2,Y_2)$ be pairs of random variables such that $X_1|_{Y_1=y}$ has the same distribution as $X_2|_{Y_2=y}$ for all $y$. Then $T(X_1,Y_1||X_2,Y_2)=T(Y_1||Y_2).$ 
\end{proposition}
\begin{proof}
By post-processing (Proposition~\ref{prop:post-processing}), $T(X_1,Y_1||X_2,Y_2)\preceq T(Y_1||Y_2).$ Let $X(y)$ be a random variable which has the distribution of $X_1|_{Y_1=y}$ and $X_2|_{Y_2=y}$. Let $M(y)=(X(y),y).$ Then $M(Y_1)=(X_1,Y_1)$ and $M(Y_2)=(X_2,Y_2)$. Therefore, by post-processing (Proposition~\ref{prop:post-processing}), we have the inequality in the other direction.
\end{proof}

\begin{proposition}[Composition~\cite{DongRS19}]
\label{prop:composition}
	Let $X_1,X_2,\dots,X_m$ be independent random variables and let $Y_1,Y_2,\dots,Y_m$ be independent random variables. Then $$T(X_1,X_2,\dots,X_m||Y_1,Y_2,\dots,Y_m)=T(X_1||Y_1)\otimes T(X_2||Y_2)\otimes \dots \otimes T(X_m||Y_m)$$ where $\otimes$ is a commutative, associative operation on functions from $[0,1]\to [0,1].$
\end{proposition}

For any random variable $X$, we have $T(X||X)\equiv \Id$ where $\Id:[0,1]\to [0,1]$ is defined as $\Id(\alpha)=1-\alpha.$ The function $\Id$ is identity for $\otimes$ operation i.e. $f\otimes \Id =f$ for all $f.$






We will the need the following proposition which explains how subsampling affects the privacy curve.
\begin{proposition}[\cite{DongRS19}]
	\label{prop:mixture_f_p}
	Let $p\in (0,1)$ and let $f=T(P||Q)$. Then $T(P|| (1-p) P + p Q)=p\cdot f + (1-p)\cdot \Id$.
\end{proposition}



Though Eqn~(\ref{eqn:tradeoff_curve}) defining the tradeoff curve requires us to take an infimum over a large class of tests, Neyman-Pearson lemma gives a very simple form for the optimal tests to use.
\begin{proposition}[Neyman-Pearson lemma]
\label{prop:Neyman_Pearson}
	Let $P,Q$ be two continuous distributions over some domain $X.$ The type I error vs type II error tradeoff function between $P$ and $Q$ is attained by the Neyman-Pearson tests which are a single parameter family of tests of the form $\phi_t:X\to [0,1]$ defined as:
	\begin{align*}
		\phi_t(x)=\begin{cases}
			1 & \text{if } Q(x) \ge t P(x)\\
			0 & \text{if } Q(x) < t P(x).
		\end{cases}
	\end{align*}
\end{proposition}

Using the Neyman-Pearson lemma, we will prove the following lemma which is crucial for our privacy analysis.
\begin{lemma}
   \label{lem:privacy_curve_knownmean_Gaussian}
	Let $Z$ be some random variable and let $f=T(Z,N(Z,1)||Z,N(0,1))$. Then $f(\alpha(t))=\beta(t)$ where:
	$$\alpha(t)=\E_Z\left[\Phi\left(-\frac{t}{Z}-\frac{Z}{2}\right)\right],\ \beta(t)=\E_Z\left[\Phi\left(\frac{t}{Z}-\frac{Z}{2}\right)\right]$$ and $\Phi(\cdot)$ is the CDF of standard Gaussian. 
\end{lemma}
\begin{proof}
Denote $P:=Z\times N(0,1), Q:=Z\times N(Z,1)$. From Neyman-Pearson lemma (Proposition~\ref{prop:Neyman_Pearson}), the type I/II errors are
\begin{align*}
\alpha(t)=&\Pr_{(z,x)\sim P}\left[\frac{Q(z,x)}{P(z,x)}\geq e^t\right]\\
=&\Pr_{z\sim Z,x\sim\mathcal{N}(0,1)}\left[\exp\left(\frac{x^2}{2}-\frac{(x-z)^2}{2}\right)\geq e^t\right]
\\
=&\Pr_{z\sim Z,x\sim\mathcal{N}(0,1)}\left[\frac{x^2}{2}-\frac{(x-z)^2}{2}\geq t\right]
\\
=&\Pr_{z\sim Z,x\sim\mathcal{N}(0,1)}\left[zx-z^2/2-t\geq 0\right]
\\
=&\E_Z\left[\Phi\left(-\frac{t}{Z}-\frac{Z}{2}\right)\right]\end{align*}
and
\begin{align*}
\beta(t)=&\Pr_{(z,x)\sim Q}\left[\frac{Q(z,x)}{P(z,x)}<e^t\right]\\
=&\Pr_{z\sim Z,x\sim\mathcal{N}(z,1)}\left[\exp\left(\frac{x^2}{2}-\frac{(x-z)^2}{2}\right)<e^t\right]
\\
=&\Pr_{z\sim Z,x\sim\mathcal{N}(z,1)}\left[\frac{x^2}{2}-\frac{(x-z)^2}{2}<t\right]\\
=&\Pr_{z\sim Z,x\sim\mathcal{N}(0,1)}\left[\frac{(x+z)^2}{2}-\frac{x^2}{2}<t\right]
\\
=&\Pr_{z\sim Z,x\sim\mathcal{N}(0,1)}\left[zx+z^2/2-t<0\right]
\\
=&\E_Z\left[\Phi\left(\frac{t}{Z}-\frac{Z}{2}\right)\right].
\end{align*}
\end{proof}

We will now prove an other key lemma which is useful for our privacy analysis.
\begin{lemma}
	\label{lem:Comparing_Gaussian_coupling}
	Let $Z,\tZ$ be two random variables such that there exists some coupling $(Z,\tZ)$ with $Z\ge \tZ\ge 0.$ Then $T(Z,N(Z,1)||Z,N(0,1))\preceq T(\tZ,N(\tZ,1)||\tZ,N(0,1))$.
\end{lemma}
\begin{proof}
	We will prove this by post-processing (Proposition~\ref{prop:post-processing}). Define a randomized map $$M(z,a)=\left(\tz,\left(\frac{\tz}{z}\right)a+N\left(0,1-\left(\frac{\tz}{z}\right)^2\right)\right)$$ where $\tz \sim \tZ|_{Z=z}$, i.e., $\tz$ is sampled from the conditional distribution of $\tZ$ given $Z=z.$ Note that $0\le\tz\le z$ because the coupling $(Z,\tZ)$ satisfies $0\le \tZ \le Z.$ Now it is easy to verify that $M(Z,N(Z,1))=(\tZ,N(\tZ,1))$ and $M(Z,N(0,1)) = (\tZ,N(0,1)).$
\end{proof}



\subsection{Proof of privacy for Algorithm~\ref{alg:DP-SGD}}
\begin{algorithm}[!h]
 \caption{Subroutine of Algorithm~\ref{alg:DPSGD_JL}}
 \label{alg:M}
 
 \begin{algorithmic}
   \State {\bfseries Input:} Vectors $\{g_1,g_2,\dots,g_N\}\subset \R^d$, clipping norm $C$, noise scale $\sigma$, number of JL projections $r$.
\\
\State Sample $v_1,v_2,\dots,v_r \leftarrow \cN(0,I_d)$
\Comment{JL projections to estimate per-sample gradient norm}
   \State For $i\in [N]$, set $M_i = \sqrt{\frac{1}{r}\sum_{j=1}^r \inpro{g_i}{v_j}^2}$
   \Comment $M_i$ is an estimate for $\norm{g_i}_2$
   \State $\tg \leftarrow  \left(\sum_{i=1}^N \min\{1,\frac{C}{M_i}\}\cdot g_i \right)+ \sigma\cdot C \cdot \cN(0,I_d)$
   \State 
   \State {\bfseries Output:} $\tg$
\end{algorithmic}
\end{algorithm}

We will first analyze the privacy of a crucial subroutine used in Algorithm~\ref{alg:DP-SGD} which is shown in Algorithm~\ref{alg:M}.
\begin{lemma}
	\label{lem:subroutine_privacy}
	Algorithm~\ref{alg:M} is $f$-DP with $$f=T(Z_r,\cN(Z_r/\sigma,1)||Z_r,\cN(0,1))$$ where $Z_r=\frac{1}{\sqrt{\frac{1}{r}\chi^2_r}}.$ Moreover $f$ can be parametrized as $f(\alpha(t))=\beta(t)$ for $t\in \R$ where:
	$$\alpha(t)=\E_{Z_r}\left[\Phi\left(-\frac{t\sigma}{Z_r}-\frac{Z_r}{2\sigma}\right)\right],\ \beta(t)=\E_{Z_r}\left[\Phi\left(\frac{t\sigma}{Z_r}-\frac{Z_r}{2\sigma}\right)\right].$$
\end{lemma}
\begin{proof}
	Let $X$ be the output of Algorithm~\ref{alg:M} with input $\{g_0,g_1,\dots,g_N\}$ and let $Y$ be the output of Algorithm~\ref{alg:M} with input $\{g_1,\dots,g_N\}$. We want to show that $T(X||Y)\succeq f$. We have $$X= \left(\sum_{i=0}^N \min\left\{1,\frac{C}{M_i}\right\}\cdot g_i \right)+ \sigma\cdot C \cdot \cN(0,I_d),\ \ Y=\left(\sum_{i=1}^N \min\left\{1,\frac{C}{M_i}\right\}\cdot g_i \right)+ \sigma\cdot C\cdot \cN(0,I_d).$$ By post-processing property (Proposition~\ref{prop:post-processing}), we have $$T(X||Y)\succeq T(v_1,v_2,\dots,v_r,M_0,X'||v_1,v_2,\dots,v_r,M_0,Y')$$ where  $$X'=  \min\left\{1,\frac{C}{M_0}\right\}\cdot g_0 + \sigma\cdot C\cdot \cN(0,I_d),\ \ Y'= \sigma\cdot C \cdot \cN(0,I_d).$$
	By Proposition~\ref{prop:sufficient-statistic}, $$T(v_1,v_2,\dots,v_r,M_0,X'||v_1,v_2,\dots,v_r,M_0,Y')=T(M_0,X'||M_0,Y').$$

	Let $U$ be a rotation matrix which rotates $g_0$ to $\norm{g_0}_2\cdot e_1\in \R^d$ where $e_1=(1,0,0,\dots,0).$ Let $X''=UX'$ and $Y''=UY'$. Since $U$ is a fixed bijective map, $$T(M_0,X'||M_0,Y')=T(M_0,X''||M_0,Y'').$$ Because of rotation invariance, $U\cdot \cN(0,I_d)$ has the same distribution as $\cN(0,I_d).$ So,
	$$X''=  \min\left\{1,\frac{C}{M_0}\right\}\cdot \norm{g_0}_2 e_1 + \sigma\cdot C \cdot \cN(0,I_d),\ Y''= \sigma\cdot C\cdot \cN(0,I_d).$$ The coordinates $(X''_i)_{i \ge 2}$ are independent of each other and $M_0,X_1''$. Similarly the coordinates $(Y''_i)_{i\ge 2}$ are also independent of each other and $M_0,Y_1''$. Moreover $X''_i$ and $Y''_i$ has the same distribution for $i\ge 2.$ Therefore by Proposition~\ref{prop:composition},
	$$T(M_0,X''||M_0,Y'')=T(M_0,X''_1||M_0,Y''_1)\otimes \Id \otimes \Id \otimes \dots \otimes \Id= T(M_0,X_1''||M_0,Y_1''),$$
	$$\text{where } X''_1=\min\left\{1,\frac{C}{M_0}\right\}\cdot \norm{g_0}_2 + \sigma\cdot C \cdot \cN(0,1) \text{ and } Y''_1=\sigma\cdot C \cdot \cN(0,1).$$ Let $\phi(M_0)=\min\left\{\frac{\norm{g_0}_2}{\sigma C}, \frac{\norm{g_0}_2}{\sigma M_0}\right\}$. We can further simplify this using Proposition~\ref{prop:sufficient-statistic} as:
	\begin{align*}
	&T(M_0,X_1''||M_0,Y_1'')\\
	&=T\left(M_0,\phi(M_0) + \cN(0,1)\ ||\ M_0,\cN(0,1)\right)\\
	&=T\left(M_0,\phi(M_0),\phi(M_0) + \cN(0,1)\ ||\ M_0, \phi(M_0),\cN(0,1)\right)\\
	&=T\left(\phi(M_0), \phi(M_0)+ \cN(0,1)\ ||\ \phi(M_0),\cN(0,1)\right)
\end{align*}
We can simplify $\phi(M_0)$ further by using the fact that $\frac{\norm{g_0}}{M_0}=\norm{g_0}_2/\left(\sqrt{\frac{1}{r}\sum_{j=1}^r \inpro{g_0}{v_j}^2}\right)$ has the same distribution as $1/\sqrt{\frac{1}{r}\chi^2_r}$. Therefore $\phi(M_0)$ has the same distribution as $$\tilde{Z}=\min\left\{\frac{\norm{g_0}_2}{\sigma C},\frac{1}{\sigma \sqrt{\frac{1}{r}\chi^2_r}}\right\}.$$ 
Let $Z_r=\frac{1}{\sqrt{\frac{1}{r}\chi^2_r}}$ and so $\tilde{Z} = \min\left\{\frac{\norm{g_0}_2}{\sigma C},\frac{Z_r}{\sigma}\right\}.$
By Lemma~\ref{lem:Comparing_Gaussian_coupling}, $$T(\tilde{Z},\cN(\tilde{Z},1)||\tilde{Z},\cN(0,1)) \succeq T(Z_r,\cN(Z_r/\sigma,1)||Z_r,\cN(0,1)).$$

Therefore, this proves that $T(X||Y) \succeq T(Z_r,\cN(Z_r/\sigma,1)||Z_r,\cN(0,1))$ where $Z_r=\frac{1}{\sqrt{\frac{1}{r}\chi^2_r}}.$ The parametrization follows from Lemma~\ref{lem:privacy_curve_knownmean_Gaussian}.
\end{proof}


\begin{theorem}
	\label{thm:final_privacy_curve}
   Let $p=B/N$ where $B$ is the batch size and $N$ is the total number of samples. Then Algorithm~\ref{alg:DPSGD_JL} is $g$-DP with $g=\min\left\{f_p^{\otimes T},(f_p^{\otimes T})^{-1}\right\}^{**}$ where $f_p=pf+(1-p)\Id$ and $f=T(Z_r,\cN(Z_r/\sigma,1)||Z_r,\cN(0,1))$.\footnote{$h^{**}$ is the double convex conjugate of $h$ (i.e., the greatest convex lower bound for $h$).}
\end{theorem}
\begin{proof}
	Algorithm~\ref{alg:DPSGD_JL} can be thought of as adaptive composition of $T$ iterations of Algorithm~\ref{alg:M}, but where the inputs to the Algorithm~\ref{alg:M} in each iteration is subsampled from the entire input with sampling probability $p=B/N$. And we already showed in Lemma~\ref{lem:subroutine_privacy} that Algorithm~\ref{alg:M} satisfies $f$-DP with $f$ as claimed. The rest of the proof is very similar to Theorem 3 in~\cite{BuDLS19} which itself builds on a similar theorem in~\cite{DongRS19}. It proceeds by applying Proposition~\ref{prop:mixture_f_p} to understand the effect of subsampling and an adaptive version of the composition in Proposition~\ref{prop:composition} to compose the privacy curves in all the $T$ iterations.
\end{proof}

One could hope to use the central limit theorem for composition from~\cite{DongRS19,BuDLS19} to find an approximate closed form expression for the final privacy curve. Unfortunately, these central limit theorems do not apply in our setting.\footnote{$\chi^2(f)$ diverges} Instead, we numerically compute the final privacy curve obtained in Theorem~\ref{thm:final_privacy_curve} making use of Lemma~\ref{lem:privacy_curve_knownmean_Gaussian}.


\subsection{Effect of JL dimension on privacy}
\label{sec:JLdim_privacy}
Since the per-sample gradient norm estimations get more accurate with JL dimension, it is clear that the privacy of DP-SGD-JL should converge to that of DP-SGD-Vanilla for large JL dimension. We also observe that privacy parameter $\epsilon$ is monotonically decreasing with increasing JL dimension and eventually converges to the $\epsilon$ for DP-SGD-Vanilla. This can be see from Figure~\ref{fig:RNNprivacy}.

\section*{Acknowledgements}
We thank Sergey Yekhanin for his constant support and encouragement during this work. We also thank Lukas Wutschitz and Osman Ramadan for helpful discussions. Finally, we thank the amazing open source community of TensorFlow for their quick response in fixing bugs which was crucial for our experiments (\cite{tfissue43449}).
\bibliographystyle{unsrtnat}
\bibliography{references}

\appendix

\section{DP-SGD and DP-Adam-JL}
\label{sec:DP-Adam-JL}
For completeness, we provide pseudo-code for DP-SGD and DP-Adam-JL used in our experiments. DP-Adam-JL satisfies exactly the same privacy bounds as DP-SGD-JL.

\begin{algorithm}[!h]
	\caption{Differentially private Adam using JL projections (DP-Adam-JL)}
	\label{alg:DP-Adam-JL}
	\begin{algorithmic}
		\State {\bfseries Input:} Examples $\{x_1,x_2,\dots,x_N\}$, loss function $\cL(\theta)=\E_{i\in [N]}[\cL(\theta;x_i)]$, initialization $\theta_0$.\\
		 Parameters: number of iterations $T$, momentum parameters $(\beta_1,\beta_2)$, noise scale $\sigma$, batch size $B$, clipping norms $(C_1,C_2,\dots,C_T)$, number of JL projections $r$.
		\State
		\For{$t=1$ to $T$}
		\State Sample $S_t=\{X_1,X_2,\dots,X_B\}\subset \{x_1,x_2,\dots,x_N\}$ uniformly at random.
		\State Define $F(\theta)= \left(\cL(\theta;X_1),\cL(\theta;X_2),\dots,\cL(\theta;X_B)\right)$
		\\
		\State Sample $v_1,v_2,\dots,v_r \leftarrow \cN(0,I_d)$ (where $\theta\in\R^d$)
		\Comment{JL projections to estimate per-sample gradient norm}
		
		\For{$j=1$ to $r$}
		\State  $(P_{1j},P_{2j},\dots,P_{Bj}) \leftarrow \grad_\theta F(\theta_{t-1}) \cdot v_j$ (using \jvp)
		\Comment{Note that $P_{ij}=\inpro{\grad_\theta \cL(\theta_{t-1};X_i)}{v_j}$}
		\EndFor
		\For{$i\in B_t$}
		\State $M_i \gets \sqrt{\frac{1}{r}\sum_{j=1}^r P_{ij}^2}$
		\Comment{$M_i$ is an estimate for $\norm{\grad_\theta \cL(\theta_{t-1};X_i)}_2.$}
		\EndFor

		\State Define $\widetilde{\cL}(\theta)=\frac{1}{B}\sum_{i\in B} \min\{1,\frac{C_t}{M_i}\}\cdot \cL(\theta;X_i).$
 \Comment{Scale the losses to clip per-sample gradients}
		
	   \State
	   \State $\tg_t \leftarrow \grad_\theta \widetilde{\cL}(\theta_{t-1}) + \frac{\sigma \cdot C_t}{B}\cdot \cN(0,I_d)$
   \Comment{Add noise to the average of clipped gradients}
	   \State
		\State $m_{t} \leftarrow \beta_{1} m_{t-1}+\left(1-\beta_{1}\right) \tilde{g}_{t} $
		\State	$ u_{t} \leftarrow \beta_{2} u_{t-1}+\left(1-\beta_{2}\right)\tilde{g}_{t}^2 $
		\State $ \eta_{t} \leftarrow m_{t} /(\sqrt{u_{t}}+10^{-8}) $
		\State Update $\theta_t\leftarrow \theta_{t-1} - \eta_t \tg_t$	
		\EndFor
		\State {\bfseries Output:} $\theta_0,\theta_1,\theta_2,\dots,\theta_T$
	\end{algorithmic}
\end{algorithm}

 \begin{algorithm}[!h]
 	\caption{Differentially private SGD (DP-SGD) from~\cite{abadi2016deep}}
 	\label{alg:DP-SGD}
 	\begin{algorithmic}
 		\State {\bfseries Input:} Examples $\{x_1,x_2,\dots,x_N\}$, loss function $\cL(\theta)=\E_{i\in [N]}[\cL(\theta;x_i)]$, initialization $\theta_0$.\\
    	Parameters: number of iterations $T$, learning rates $(\eta_1,\eta_2,\dots,\eta_T)$, noise scale $\sigma$, batch size $B$, clipping norms $(C_1,C_2,\dots,C_T)$.
 		\State
 		\For{$t=1$ to $T$}
 		\State Sample $S_t=\{X_1,X_2,\dots,X_B\}\subset \{x_1,x_2,\dots,x_N\}$ uniformly at random.
 		\For{$i=1$ to $B$}
 		\State $\hat{g}_i\gets\min\{1,\frac{C_t}{\|\nabla_\theta\cL(\theta_{t-1};X_i)\|_2}\}\cdot \nabla_\theta\cL(\theta_{t-1};X_i)$
 		\Comment{Clip the per-sample gradients to $\ell_2$-norm at most $C_t$}
 		\State $\tg_t \leftarrow \frac{1}{B}\left(\sum_{i\in B_t}\hat{g}_i \right)+ \frac{\sigma C_t}{B}\cdot \cN(0,I_d)$
 		\Comment{Average the clipped gradients and add noise}
 		\State Update $\theta_t\leftarrow \theta_{t-1} - \eta_t \tg_t$
 		\EndFor
 		\EndFor
 		\State {\bfseries Output:} $\theta_0,\theta_1,\dots,\theta_T$
 	\end{algorithmic}
 \end{algorithm}

\end{document}